\documentclass[10pt,twocolumn,letterpaper]{article}

\usepackage{wacv}
\usepackage{times}
\usepackage{epsfig}
\usepackage{graphicx}
\usepackage{amsmath}
\usepackage{amssymb}

\def\wacvPaperID{1} 

\wacvfinalcopy 

\ifwacvfinal
\def\assignedStartPage{1} 
\fi

\ifwacvfinal
\usepackage[breaklinks=true,bookmarks=false]{hyperref}
\else
\usepackage[pagebackref=true,breaklinks=true,colorlinks,bookmarks=false]{hyperref}
\fi

\ifwacvfinal
\else
\fi

\begin{document}

\title{ Deep Generative Framework for Interactive 3D Terrain Authoring and Manipulation}

\author{Shanthika Naik 
	\quad Aryamaan Jain
	\quad Avinash Sharma
	\quad KS Rajan \\
	IIIT Hyderabad
}


\maketitle

\begin{abstract}
   Automated generation and (user) authoring of the realistic virtual terrain is most sought for by the multimedia applications like VR models and gaming. The most common representation adopted for terrain is {\emph {Digital Elevation Model}} (DEM). Existing terrain authoring and modelling techniques have addressed some of these and can be broadly categorised as: \textit{procedural modeling, simulation method}, and  \textit{example-based methods}. In this paper, we propose a novel realistic terrain authoring framework powered by a combination of VAE and generative conditional GAN  model. Our framework is an example-based method that attempt to overcome the limitations of existing methods by learning a latent space from real world terrain dataset. This latent space allows us to generate multiple variants of terrain from a single input as well as interpolate between terrains, while keeping the generated terrains close to real world data distribution. We also developed an interactive tool, that lets the user generate diverse terrains with minimalist inputs. We perform thorough qualitative and quantitative analysis and provide comparison with other SOTA methods. We intend to release our code/tool to academic community. 
\end{abstract}

\section{Introduction}
%
Terrain modelling aims to create a digital representation of the real world topography and is useful in both scientific applications of land surface processes like flooding, soil erosion as well as virtual terrain rendering in graphics and computer vision applications.
\begin{figure} [!ht]
\centering
\includegraphics[width=0.4\textwidth]{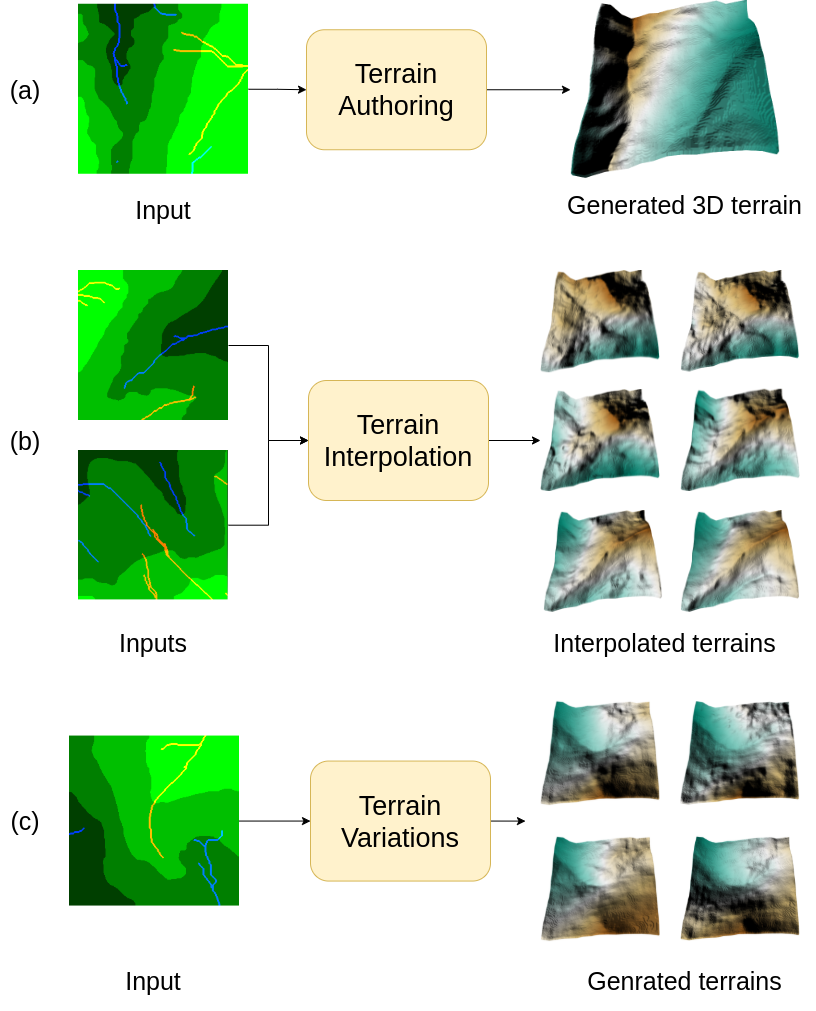}
\caption{The proposed framework is useful across multiple use cases, namely, interactive terrain authoring, terrain interpolation and automated generation of terrain variations.}
\label{samples}
\end{figure}
\begin{figure*}[!ht]
\centering
\includegraphics[width=\textwidth]{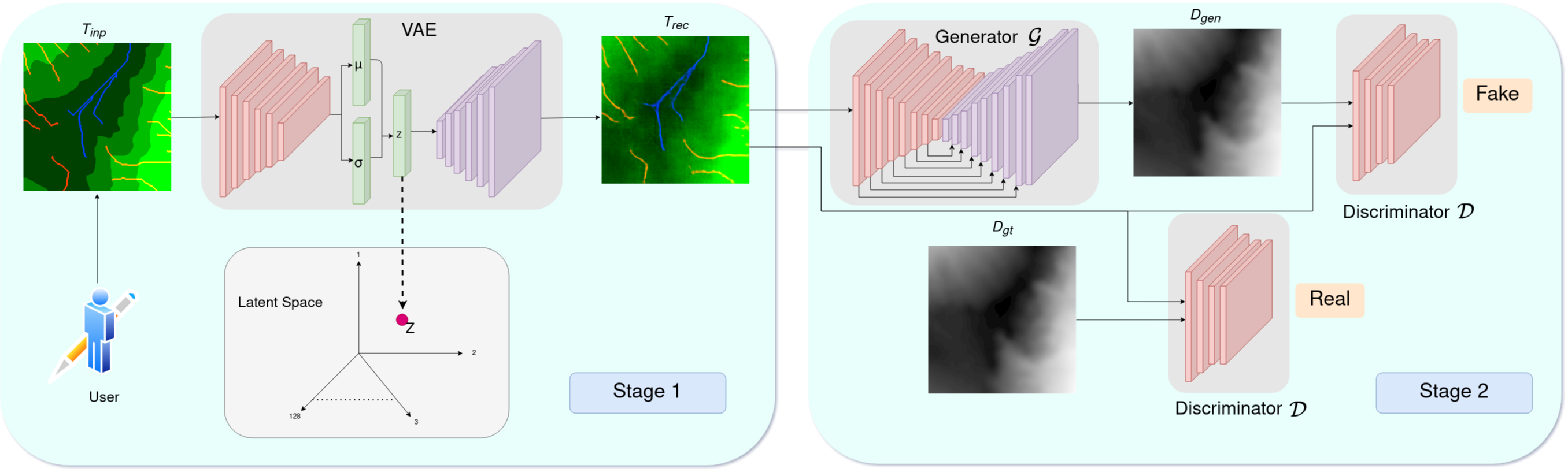}
\caption{The proposed architecture is composed of two stages. \textbf{Stage 1:} For each input $T_{inp}$, the VAE learn a generative latent space while reconstructing the topographic map $T_{rec}$. \textbf{Stage 2:} conditional GAN take generated topographic map $T_{rec}$ as input and learn to generate realistic terrain DEM in a supervised learning setup.}
\label{fig:arch-diag}
\end{figure*}
In particular, automated generation and (user) authoring of the realistic virtual terrain is most sought for by the multimedia applications like Virtual Reality (VR) models and gaming. 
The real world terrains undergo a range of natural transformations such as erosion, weathering, and landslides over the years. This leads to various complex topographies such as hills, mountain ranges, canyons, plateaus, and plains. Some of these significant and unique terrain features are apparent at different scales. All these contribute to making the terrain generation and authoring a challenging task. The most common representation adopted for terrain is {\emph {Digital Elevation Model}} (DEM). DEM is a discrete grid representation of the ground elevation and its quality is dependent on the resolution or scale of the grid or pixel. 

Existing terrain authoring and modelling techniques have addressed some of these and can be broadly categorised as: \textit{procedural modeling, simulation method}, and \textit{example-based methods} \protect{\cite{rev_terrain_modelling}}. Procedural modeling use algorithms to reproduce the effects of physical phenomena. Simulation methods perform  simulations of real-world phenomena such as thermal and hydraulic erosion, weathering and land formation where the modelling of the processes and its interactions involving physical elements lead to the formation of the targeted terrains. Example-based methods use information extracted from real world examples. This method provides better user control over the generated outputs.

Recent advancements in deep learning enabled the ability to learn diverse terrain features for tasks like terrain amplification~\cite{kubade}, modifications~\cite{landslide}, etc. In the context of deep learning based automated terrain authoring, the literature is very sparse. One of the most relevant example-based terrain authoring method proposed trains a conditional Generative Adversarial Network (cGAN), referred in this work as {\emph TSynthNet}~\cite{base_paper}, on large set of real world terrain data to generate realistic virtual terrains from hand drawn user inputs. However they provide a limited user control and generate a single terrain for given input. Additionally, their method allows the use of either drawings of level-sets or ridge/valley strokes (with altitude cues). However, these two representations are complementary and jointly provide richer information for terrain authoring task. One major limitation of their method is that they fail to learn a generative latent space and hence are unable to perform terrain manipulation tasks. 

In this paper, we propose a novel realistic terrain authoring framework powered by a combination of Variational Auto-encoder (VAE), and conditional GAN model. Our framework attempt to overcome the limitations of TSynthNet~\cite{base_paper} by learning a latent space from real world terrain dataset. This latent space allows us to generate multiple variants of terrain from a single input as well as interpolate between terrains, while keeping the generated terrains close to real world data distribution. 
We also developed an interactive tool, that lets the user model diverse terrains with minimalist inputs. We perform thorough qualitative and quantitative analysis and provide comparison with TSynthNet~\cite{base_paper} to show superiority of our method over SOTA. 
We summarise our contributions as follows: 
\begin{itemize}
\item We propose a novel terrain authoring framework, powered by VAE and cGAN and provide thorough empirically evaluation for the same on real world dataset.
\item We propose to learn a generative latent space for topographic maps using VAE that enables terrain interpolation and automated variant terrain generation.
\item We design a novel VAE loss function to exploit sparse topographic features like ridge/valley lines.   
\end{itemize}


\textit{Example-based methods} use real world examples or user defined sketches for terrain synthesis. Texture-based terrain synthesis technique takes a set of input images and produces an output with similar characteristics by deriving from common features in input data. \cite{exmp1} is a texture-based method that takes user constraints in the form of points, curves and painted terrains. \cite{exmp2} lets user edit existing terrains by providing sketch silhouettes. \cite{exmp3} combines interactive and procedural methods to let user render complex waterfall scenes. They also allow user to edit existing terrains to model waterfalls. Sketch-based methods are often tedious as it requires manual editing whereas example-based methods are limited by dataset and cannot generate new ones unless present in the input. 


\textit{Generative methods in deep learning} attempts task of generating unique, diverse data. \cite{boltz} is one of the earliest unsupervised generative model. \cite{autoencoder} use convolution neural networks (CNNs) to compress input data into a low dimensional latent vector and generate images from these vectors. VAE \cite{vae} maps the  input data on to a probabilistic distribution, by learning their parameters for each input. They manage to model the entire dataset on to latent space. The data points similar to each other lie close on this learnt latent space. It is possible to sample data similar to each other by sampling from the learnt distribution. One can also traverse this latent space to interpolate between two data points. GANs \cite{gan} use a combination of generative and adversarial networks for training. The adversarial loss used makes sure that the generated sample distribution and true data distribution are close to each other. \cite{dcgan} use CNNs with GANs to generate unique images from vectors sampled from a known distribution. Conditional GANs (cGAN) \cite{cGAN} use auxiliary information along with input to to have better control over generated output. Pix2pix \cite{pix2pix} is a cGAN which performs image to image translation. They also provide an application of translating hand drawn input into meaningful image. \cite{deep_face, mask_face} are a face editing models that first learn a local feature embedding for each component of face, then further user generative network to generate images with incorporated editing. Learning a latent space gives more flexibility for authoring and editing applications.
TSynthNet~\cite{base_paper} is one of the first to use a deep learning model for example based terrain synthesis. They use a cGAN model borrowed from \cite{pix2pix} to generate terrains from hand drawn topographic maps (either level-sets or ridge/valley lines). 

Please refer to supplementary material for detailed discussion on other related works in the literature. 

\section{Methodology}
We propose a realistic virtual terrain authoring framework for generating terrains in the form of Digital Elevation Models (DEMs) from hand drawn sketches. Our two-stage framework attempt to learn the topographical structure of real-world terrains from existing datasets and generate plausible DEM from input sketches that can be thought of as {\it  topographic maps} primarily consisting of DEM level-sets and ridge/valley lines representing underlying abstract topological features of the terrain.
In the first stage, we aim to learn a generative latent space for topographic maps using a Variational Auto-encoder (VAE)~\cite{vae} model from real world terrain dataset. We extract ground truth topographic maps from real world terrain data (see Section~\ref{sec:dataset}) and autoregress using VAE to learn the latent space. Learning such a latent space enables two key use-cases of proposed method, namely, automated generation of multiple variants of terrain from a single user input sketch, and interpolate between two terrains, while keeping the generated virtual terrains closer to real world dataset consisting of realistic topographical features. The output of this stage is the latent topographic map depicting generated level-sets and ridge/valley lines. Subsequently, this output is fed to the second stage which consists of conditional Generative Adversarial Network (cGAN) (Pix2pix~\cite{pix2pix}) model that generates plausible DEM output. Although our second stage is inspired from TSynthNet~\cite{base_paper}, we propose to combine DEM level-sets and ridge/valley line strokes into a single topographic map as these two representations are complementary and jointly provide richer information for terrain authoring task. We propose a novel VAE loss to emphasize the sparse ridge/valley lines in the joint topographic map representation. \autoref{fig:arch-diag} show the overview of proposed two-stage framework. 


%


\subsection{Stage 1: Latent Space Learning}

Let $T_{inp}$ be our (hand drawn) input sketches representing a rough topographic map. We assume that these inputs belong to some parametric distribution $p_{\theta}(T_{inp}|z)$ and each terrain is generated by some random process, involving an unobserved random variable $z$. Calculating $p_{\theta}(T_{inp}|z)$ is hard as $p(T_{inp})$ is intractable. Hence, we use the VAE to approximate another distribution $q(z)$ such that it is similar to learns the distribution parameters $\mu$ and $\sigma$, from which the latent vector $z$ is sampled. We use a re-parameterization trick to sample $z$ as $z = \mu + \sigma *\epsilon$, where $\epsilon$ is sampled from a Standard Normal distribution. This sampled vector is fed to the decoder which predicts $T_{rec}$, that is the reconstruction of original input $T_{inp}$.
\begin{equation}
\label{eq:BCE_loss}
L_{BCE}(T_{inp},T_{rec}) = - [ T_{inp} \log{T_{rec}} + (1-T_{inp}) \log{(1-T_{rec})} ]
\end{equation}

\begin{equation}
\label{eq:recons_loss}
\begin{split}
L_{recons} = L_{BCE}(T_{rec}^g,T_{inp}^g) + \alpha [L_{BCE}(T_{rec}^r,T_{inp}^r) \\+ L_{BCE}(T_{rec}^b,T_{inp}^b)] 
\end{split}
\end{equation}

 We propose a novel auto-regressive reconstruction loss $L_{recons}$ between VAE input $T_{inp}$ and output $T_{rec}$ by modifying the traditional Binary Cross Entropy (BCE) loss (\autoref{eq:BCE_loss}) to emphasize the ridge/valley lines in the topographic map.  Here, we represent the input topographic map $T_{inp}$ as RGB image where the Green channel ($T_{inp}^g$) is dedicated to represent level-sets while the Red  ($T_{inp}^r$) and Blue ($T_{inp}^b$) channels are used to represent ridge and valley lines, respectively. A similar representation exist for output (reconstructed) topographic map ($T_{rec}$). We propose to give higher weightage to the loss on red and blue channels so as to give more importance to red and blue channels representing sparse ridge and valley lines/strokes in the topographic map sketches. This is desired in our joint representation of topographic map as ridge/valley lines/strokes are relatively sparse (occupying fewer pixels) in comparison to dense level-sets maps (green channel). This helps the model learn sharp features of river valleys and hill ridges. 
 Additionally, a traditional KL divergence loss $L_{KL}$  (\autoref{eq:KL_loss}) ensure  that the probability distribution of latent vector $z$ follows a Standard Normal distribution. 
 \begin{equation}
\label{eq:KL_loss}
L_{KL} = \sum_{i=1}^{d}{[\sigma^2 + \mu^2 - \log{\sigma^2}]} 
\end{equation}

 Thus, the final VAE loss $L_{VAE}$ (\autoref{eq:vae_loss}) is a combination of reconstruction $L_{recons}$ (\autoref{eq:recons_loss}) and  KL divergence loss $L_{KL}$ (\autoref{eq:KL_loss}). The $\gamma$ parameter in \autoref{eq:vae_loss} is the weighting of the latent loss $L_{KL}$ which is set to 0.65, $\alpha$ is weight of red and blue channels in reconstruction loss which is set to 5, $\mu$ and $\sigma^2$ are the mean and variance learnt by the network. 
\begin{equation}
\label{eq:vae_loss}
L_{VAE} = L_{recons} + \gamma * L_{KL}
\end{equation}
The VAE output $T_{rec}$ is a topographic map, which is further fed to our second stage for generating the plausible DEM for completion of terrain authoring task. However, apart from traditional terrain authoring task, the key importance of our multi-stage framework is that the VAE based learning of latent space also enables performing interpolation of terrains as well as generating novel variants of a given terrain, as explained below.

\subsubsection{Terrain Interpolation}
 We can use the latent space for interpolation between two terrains in the lower dimensional (latent) feature space ($z$). Given input (topographic map) sketches $T_{inp}^1$ and $T_{inp}^2$, we can obtain their corresponding latent representations in terms of means $\mu_1$ \& $\mu_2$ and standard deviations $\sigma_1$ \& $\sigma_2$. We can subsequently interpolate between the $z$ of these two inputs as $z = \gamma * z_1 + (1-\gamma) * z_2$ where $\gamma$ ($0\leq\gamma\leq1$) is an interpolation parameter defining degree of dominance of $T_{inp}^1$ over $T_{inp}^2$ in the interpolated output topographic map. The interpolated topographic map output is fed to the second stage to get the final DEM. 

\subsubsection{Generating Terrain Variants}
We can further use this learnt latent space for sampling new terrain variant for automated variant generation. Given an input $T_{inp}$, we can obtain corresponding $\mu$ and $\sigma$ vectors from the encoder. Subsequently, we can sample new $z$ vectors as $z = \mu + \sigma *\epsilon$, where we can vary $\epsilon$ parameter to get different output terrains ($T_{rec}$) with variations in terrain topographic features as compare to input $T_{inp}$. The generated topographic map variant is fed to the second stage to get the final DEM. 

\begin{figure*} [ht]
\centering
\includegraphics[width=\textwidth]{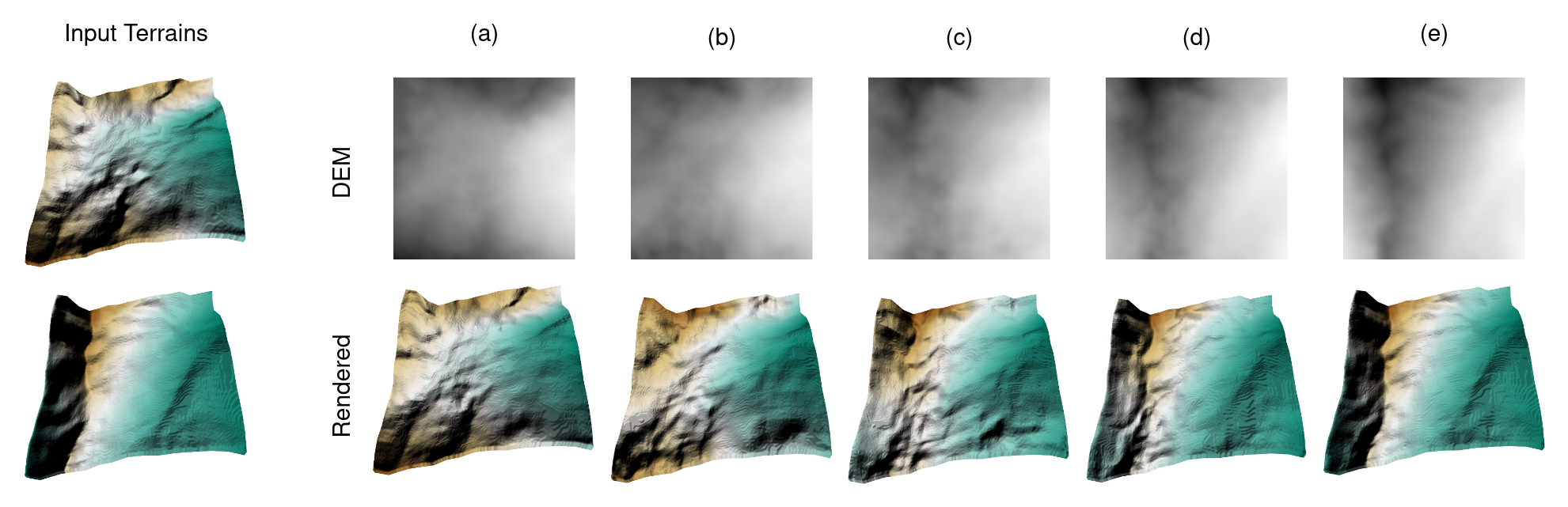}
\caption{Given two input terrains, we interpolate between the two using the latent space of VAE. We combine the two using the formula $z = \gamma * z_1 + (1-\gamma) * z_2$. The $\gamma$ values in the above figure are as follows: a) $\gamma = 0.167$ b) $\gamma = 0.334$ c) $\gamma = 0.501$ d) $\gamma = 0.668$ e) $\gamma = 0.835$. Even though in this case we show only 5 images, we can have more smooth transition by reducing the value of $\gamma$.}  
\label{fig:int_res} 
\end{figure*}
\begin{figure*} [!ht]
\centering
\includegraphics[width=\textwidth]{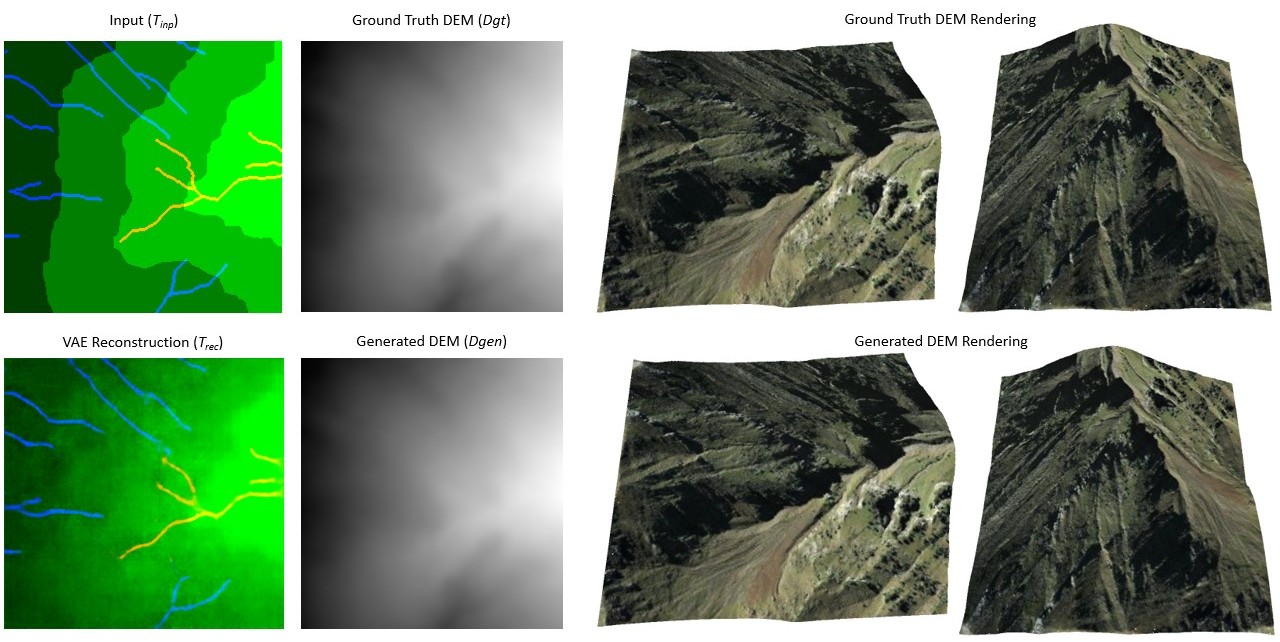}
\caption{The first column shows the input and the VAE reconstructed images. The second column gives a comparison between the ground truth and generated DEMs. The last two columns show the 3D rendering of these DEMs overlaid with the satellite image.}  
\label{fig:res1}
\end{figure*}

\begin{figure} [h!]
\centering
\includegraphics[width=0.45\textwidth]{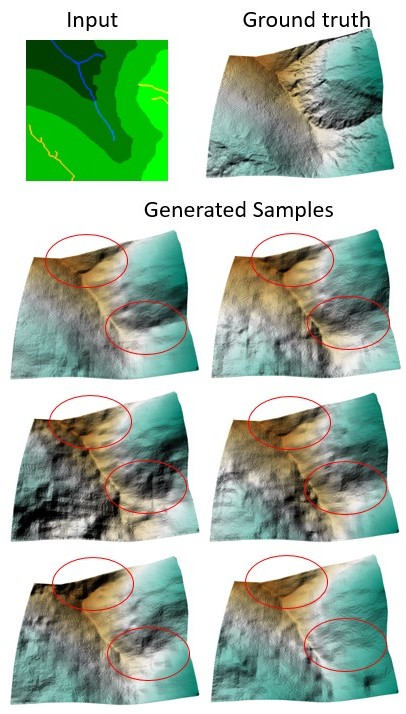}
\caption{The variants of a given terrains generated from a single input topographic map sketch. The red circled area show the clear difference between the generated variants and original terrains.}  
\label{fig:samples}
\end{figure}
\begin{figure*}[ht]
\centering
\includegraphics[width=0.8\textwidth]{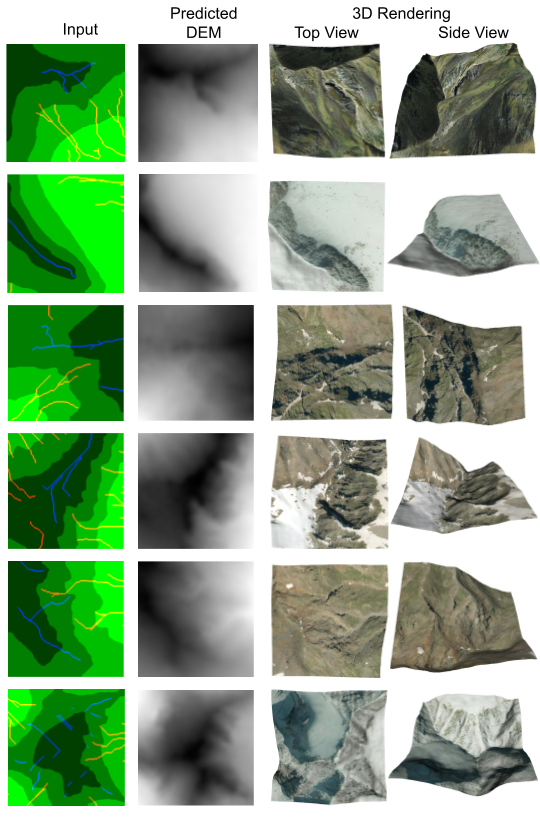}
\caption{Diverse terrains generated by our framework.}  
\label{fig:terrains}
\end{figure*}
\subsection{Stage 2: DEM Generation}
The aim of this stage is to generate the DEM given user topographic map sketch or in our case generated sketch from previous stage. Similar to TSynthNet~\cite{base_paper}, we adapt Pix2pix~\cite{pix2pix}, a conditional GAN model for this task. 
%
Traditionally, GAN~\cite{gan} are based on the idea of making a given noise distribution to match the desired data distribution. The conditional GANs \cite{cGAN} conditions both generator and discriminator on some auxiliary information to have better control over the generated output. The generator network $\mathcal{G}$ attempts to learn this mapping while the discriminator $\mathcal{D}$ learns to discriminate the generated data from ground truth data. Thus, the overall network is trained such that both generator and discriminator reach a Nash equilibrium by playing a two player minmax game while optimizing the value functions. 

Let's represent ground truth DEM as $D_{gt}$ and $D_{gen}$ as generated/predicted DEM by the Pix2pix model for the input topographic map sketch $T_{rec}$. Let $f$ be the feature embedding learnt by the generator encoder, the value function can be defined as:
%
\begin{equation}
\label{eq:value-func}
\begin{split}
    V(\mathcal{D},\mathcal{G}) = \mathbb{E}_{D_{gt}}[\;\log \mathcal{D}(D_{gt})\;] \\ + \mathbb{E}_{f} [\;\log(1 - \mathcal{D}(\mathcal{G}(f)))\;]
\end{split}
\end{equation}

Our model uses hand drawn input topographic sketch maps as an auxiliary information in order to produce plausible DEMs. We use L1 loss for reconstruction from the generator i.e., $L1(\mathcal{G})$. So the final loss for cGAN training becomes:
\begin{equation}
L = \underset{\mathcal{G}}min\; \underset{\mathcal{D}}max\; [V(\mathcal{D},\mathcal{G}) + L1(\mathcal{G})]
\end{equation}

\section{Experimental Setup}
\subsection{Dataset}   
\label{sec:dataset}

We use popular DEM dataset used by other relevant works in the literature e.g., ~\cite{argudo2018terrain,datasetpaper} which is part of publicly available high-resolution DEMs of mountain ranges named Pyrenees~\cite{icc} and Tyrol~\cite{sbg}, respectively. DEM patches with a resolution of 2m/pixel has been used as ground truth elevation maps. Original DEM tiles were split into 200x200 pixels. Instead of using the whole dataset of 22000 image patches for training and 11000 image patches for testing, we randomly sample 3000 image patches for training and 878 image patches for testing. More details about the dataset can be referred from \cite{datasetpaper}.
We prepare the training dataset by extracting the topographic map input sketches from DEMs (please refer to supplementary material for details).

\subsection{Implementation Details}

Our VAE model (in stage 1) is a 12 layer network with 6 layers in encoder and 6 layers in decoder. The latent space dimension is set to 128. All the layers of encoder consists of a 3x3 convolution with stride of 2 and padding of 1, followed by Batch Normalization and using Leaky ReLU non-linearity. Due to stride of 2, the image resolution reduces by two after each layer, and we double the number of channels in each layer starting from 32 in first and 1024 in the final layer of encoder. We then flatten the output of encoder and pass it through fully connected layer to get 128 dimension mean and log variance. This mean and log variance is then used to sample and then the decoder is used to reconstruct the input. Each decoder layer consists of Transposed Convolution with stride of 2 and input and output padding of 1, followed by batch normalization and leaky ReLU as activation. Decoder also consists of 6 layers, so we reach the same resolution as input. At the end we have a convolution layer with 3x3 kernel, stride of 1, padding of 1 and output channels 3.
Adam optimizer was used to update the parameters with learning rate of $0.001$ and an exponential scheduler with gamma set to $0.95$ while training the VAE with a batch size of $64$ for $100$ epochs.

Our conditional GAN generator (in stage 2) is a U-net inspired Pix2pix architecture~\cite{pix2pix}. 
This model was trained using Adam optimizer with learning rate of $0.0002$, $\beta_1$ set to $0.5$ and $\beta_2$ as $0.999$ for both Generator and Discriminator. We use a batch size of $32$ here. 

We used 3000 terrain patches (of 256$\times$256) for training and 878 terrain patches (of 256$\times$256) for testing both stage 1 and 2 models of our method. All our experiments were performed on a single Nvidia GTX 1080Ti. The network implementations were done using Pytorch.

\section{Results \& Discussion}
\subsection{Qualitative Evaluation}
We demonstrate the result of the proposed pipeline in Figure \ref{fig:res1}. The figure shows the comparison between rendered ground truth and predicted DEM draped with corresponding satellite images. This shows that the model is able to generate realistic terrains close to the ground truth. Some more examples of diverse terrains generated are shown in Figure~\ref{fig:terrains}. \\
\\
{\bf Generating Terrain Variants:} 
We utilise the latent space created by VAE to generate different samples from the same input. Different terrains generated from the latent space encoding of the same input topographic map are shown in Figure~\ref{fig:samples}. The red circled regions shows the variation in the generated terrains as compared to the original terrain. We can observe the generated terrains have realistic but slightly different topographical features from that of input terrain. This provides the user the flexibility to generate multiple terrain DEMs and use them for large scale generation of virtual terrain maps.\\
\\
{\bf Terrain Interpolation:} 
The latent space can also be used for automated fusion of topographic features across two terrains. Figure \ref{fig:int_res} shows an example interpolation of two input terrains in the latent space for different values of $\gamma$ parameter. This enables the user to generate new virtual terrain maps by combining a given pair of input terrain maps.

Please refer to supplementary video for 3D visualization of qualitative results. 
\subsection{Quantitative Evaluation \& Ablation Study}
\begin{table}[]
\begin{tabular}{| p{5cm} | c |}
\hline
 Method & MSE      \\
 \hline
 TSynthNet~\cite{base_paper} & 30.157 \\
 \hline
 Only VAE &   57.265 \\
 \hline
 Our model (VAE+cGAN) & {\bf 28.4024}  \\
 \hline
 Our model (VAE+cGAN) without modified VAE loss & 73.611  \\
 \hline
\end{tabular}
\label{tab:qunatitative-results}
\caption{Quantitative comparison with variant of our method and Pix2pix~\cite{pix2pix}.}
\end{table}
%
We perform quantitative analysis and compare our model with TSynthNet~\cite{base_paper}. Additionally, we also compare our method with two variants namely, ``Only VAE" and ``VAE+cGAN without modified VAE loss". We attempted to learn the DEM generation using only VAE in the former model. The latter is an ablative variant of our model without modified VAE loss. 
We train all these models with same train/test split on terrain dataset (see \autoref{sec:dataset}) and reported Mean Squared Error (MSE) as the evaluation metric. 

We provide quantitative evaluation results in Table~\ref{tab:qunatitative-results} . We can observe that we obtained superior performance with  MSE of $28.4024$ from our model (VAE+cGAN with modified VAE loss).  The ablative comparison with VAE+cGAN without modified VAE loss (i.e., MSE value of $73.67$) show that our proposed modified VAE loss that account for ridge/valley lines significantly improves the performance of our method. Only VAE model (i.e., single stage VAE based DEM generation) also performs inferior (with MSE of $57.265$) to our model justifying need of two stage framework. In comparison to TSynthNet~\cite{base_paper}, our model yield lower MSE value.   It is important to note that apart from achieving lower MSE than TSynthNet on terrain generation task, our method also enables terrain interpolation and variant generation using the learnt VAE latent space. Thus, our model yield superior quantitative performance as compared to other methods/variants. 

%
%
%

We also performed two additional ablation experiments to justify our proposed method. In the first experiment, we retain the same architecture as our method but train it differently. Here we first train the VAE of first stage but during training of cGAN in the second stage, we let the gradients back propagate all way to decoder (while VAE encoder weights are kept frozen). However, we achieved an MSE of $80.67$ quiet inferior to our model.
In the second experiment we initially train the stage 1 VAE to learn the latent space. In the second stage we again learn another VAE but with borrowing and freezing the encoder weights from first stage and train only the decoder for generating DEMs. Thus, it is still a two stage pipeline with two VAEs. We obtained an MSE of $35.67$ with this variant.
These ablative studies justifies choice of our proposed two stage framework and learning strategy. 

\subsection{User Study}

\begin{figure}[!ht]
\centering
\includegraphics[width=0.4\textwidth ]{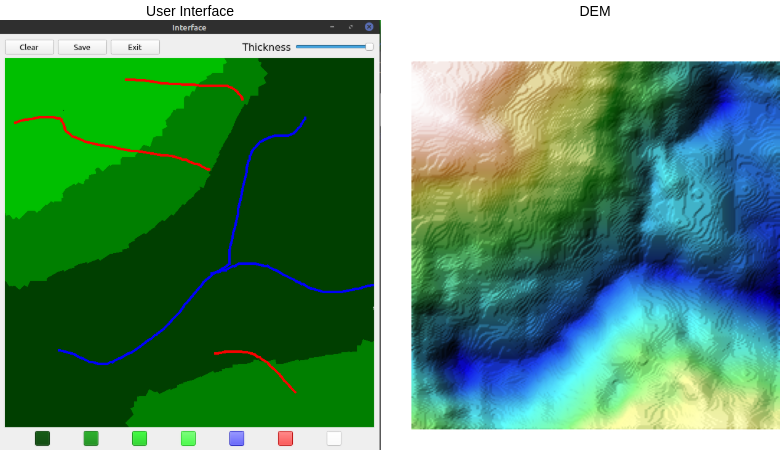}
\caption{The User Interface and the generated DEM for the input topographic map sketch. } 
\label{fig:ui}
\end{figure}

\begin{figure}[!ht]
\centering
\includegraphics[width=0.4\textwidth  ]{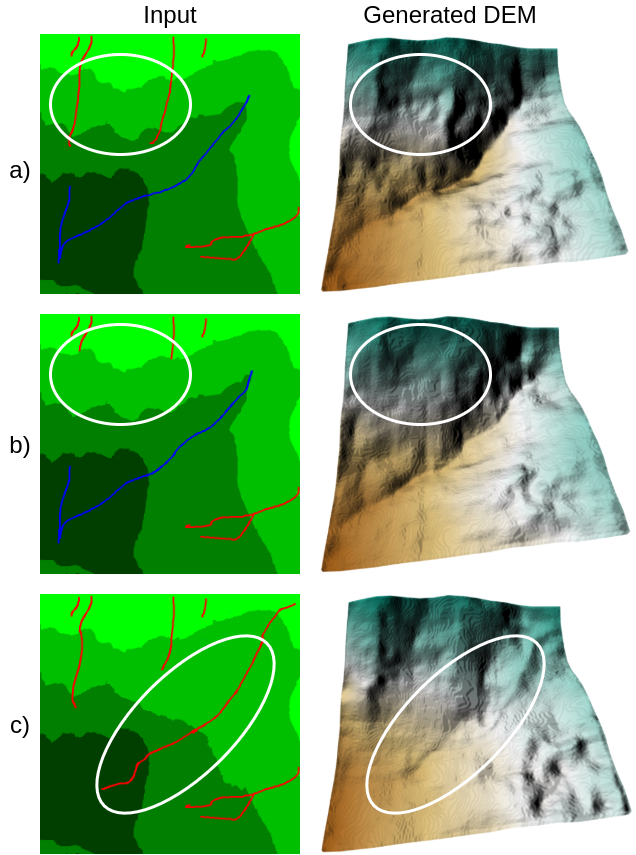}
\caption{Terrains generated during an interactive terrain editing session.  (White circles points to area of input editing).} 
\label{fig:ui2}
\end{figure}

We performed a detailed user study involving $6$ users. In order to test the degree to which our model can generate realistic terrains, we presented users (students trained in geographical information system course) with set of generated (reconstructed) and ground truth terrain pairs overlaid with satellite images, similar to ones shown in Figure~\ref{fig:terrains} and asked them to differentiate between the two. The outcome was that users were unable to decisively differentiate and choose the real terrain only $50\%$ of the time. Total $83.3\%$ of the users agreed that the terrains generated are very realistic while $16.7\%$ said that it is fairly realistic. 

In the second experiment, we provide the user with a simple interface to draw input sketches. The designed interface is simple with controls such as Clear, Save, Exit. We provide the option to vary brush thickness so that the dense level-sets can be drawn with only few strokes. The user interface and the DEM generated for a hand draw user input is shown in Figure \ref{fig:ui}. The input can also be interactively edited to get desired output. Figure~\ref{fig:ui2} shows visualization of generated terrains during an interactive terrain editing session. We asked the users several questions regarding the interface and the application such as 1) whether the input is intuitive, 2) if it is easy to express one's intent, 3) if the terrains generated follow the sketches etc. $33.3\%$ users said that the input is very intuitive while $66.7\%$ users agreed that it is fairly intuitive. $50\%$ of the user strongly agree that the generated terrain follow the input sketches, while the remaining $50\%$ fairly agree. All the users strongly agree that the system is fast and reactive. When asked to rate on a scale of 1 to 5, on how easy it was to express one's intent, $50\%$ of the users gave a rating of 5, $16.7\%$ gave a rating of 3 and $33.3\%$ gave a rating of 2. We observe that the users were able to generate DEMs with ease after a couple of attempts. 

\section{Conclusion}
We proposed a novel realistic terrain authoring framework powered by a combination of Variational Auto-encoder and  conditional GAN model. Our framework learns a generative latent space from real world terrain dataset. This latent space allows us to generate multiple variants of terrain from a single input as well as interpolate between terrains, while keeping the generated terrains close to real world data distribution. 
While a preliminary interactive tool has been developed and used here, we further intend to provide user control to generate the terrain variants and interpolated terrains. The thorough qualitative and quantitative analysis and comparison with other SOTA methods supports the superior outcome of our approach.

{\small
\bibliographystyle{ieee_fullname}
\bibliography{egpaper}
}

\end{document}